\documentclass[runningheads,a4paper]{llncs}

\usepackage{amssymb}
\usepackage{graphicx}
\usepackage{cite}
\usepackage{multirow}

\usepackage{algorithm}
\usepackage{algpseudocode}

\usepackage{url}
\urldef{\mailsa}\path|{nguyenthithanh.dang,patrick.decausmaecker}@kuleuven-kulak.be|
\newcommand{\keywords}[1]{\par\addvspace\baselineskip
\noindent\keywordname\enspace\ignorespaces#1}

\usepackage{color}

\newcommand{\lit}[1]{{\textit{#1}}}

\begin{document}

\mainmatter  

\title{Characterization of neighborhood behaviours in a multi-neighborhood local search algorithm}

\titlerunning{Characterization of neighborhood behaviours}

%
%
\author{Nguyen Thi Thanh Dang \and Patrick De Causmaecker}
\authorrunning{N.T.T.Dang, P.De Causmaecker}

\institute{KU Leuven KULAK, CODeS, iMinds-ITEC\\
\mailsa}

%
%

\toctitle{Characterization of neighborhood behaviours in a multi-neighborhood local search algorithm}
\tocauthor{N.T.T.Dang, P.De Causmaecker}
\maketitle

\begin{abstract}
We consider a multi-neighborhood local search algorithm with 
a large number of possible neighborhoods. Each neighborhood 
is accompanied by a weight value which represents the probability 
of being chosen at each iteration. These weights are fixed before 
the algorithm runs, and are considered as parameters of the algorithm. 
Given a set of instances, off-line tuning of the algorithm's parameters can be done by automated algorithm configuration tools 
(e.g., SMAC). However, the large number of neighborhoods can make the 
tuning expensive and difficult
even when the number of parameters has been reduced by some intuition. 
In this work, we propose a systematic method to characterize each neighborhood's behaviours, 
representing them as a feature vector, and using cluster analysis 
to form similar groups of neighborhoods. 
The novelty of our characterization method is the ability of reflecting changes of behaviours according to hardness of different solution quality regions. 
We show that using neighborhood clusters instead of individual neighborhoods helps to reduce the parameter 
configuration space without misleading the search of the tuning procedure. 
Moreover, this method is problem-independent and potentially can be applied in similar contexts.
\keywords{algorithm configuration, clustering, multi-neighborhood local search}
\end{abstract}

\section{Introduction}

Because optimization algorithms are usually highly parameterized, algorithm parameter tuning/configuration is an important task. 
Given a distribution of problem instances, we need to find parameter configurations that optimize a predefined performance measure over the distribution, such as mean of optimality gap. For the last fifteen years, automated algorithm configuration has been extensively studied \cite{hoos2012automated}. General-purpose automated algorithm configuration tools such as SMAC \cite{hutter2011sequential} and irace \cite{lopez2011irace} have been successfully applied in several studies. 

In this work, we consider the parameter tuning problem of a multi-neighborhood local search algorithm \cite{wauters2015winning}, 
which consists of a large number of neighborhoods. The algorithm is the winner of the Verolog Solver Challenge 2014 \cite{heid2014verolog}. 
At each iteration, a neighbor solution is generated by a randomly chosen neighborhood with a probability defined by a weight value in the range of [0,1]. 
Weights of all neighborhoods are fixed before the algorithm runs, and are considered as algorithm parameters. Given a set of six (large) instances provided by the challenge, automated algorithm configuration tools can be used to tune the algorithm parameters. 
However, the large number of parameters (28 real parameters for the weight values and 2 integer parameters for the local search) might deteriorate the tuning tool's efficiency, 
especially in our case where each run of the algorithm is not computationally cheap (600 seconds per run for each instance). 
A potential solution is to cluster neighborhoods into groups and assign a common weight value to each. 
It can help to reduce the algorithm configuration space, hoping to make use of available tuning budgets more efficiently. 
The key question raised while doing such a clustering is how to characterize each neighborhood behaviours over a set of instances and represent them as a feature vector. In this paper, we propose a method to do so. This method is problem-independent and does not depend on any specific local search. Moreover, it can be done during stages of algorithm development, e.g., testing, manual/automated tuning.

This paper is organized as follows. We describe the tuning problem in more detail in section 2. The method for characterizing neighborhoods' behaviours and clustering them is explained in section 3. Section 4 shows the advantage of using clustering in automated parameter tuning and experimental results. Finally, section 5 gives conclusion and discussion on future work.

\section{Parameter tuning for a multi-neighborhood local search algorithm}
\label{sec:problem_statement}

The algorithm considered in this work, which was developed by CODeS group's members of the University of Leuven (Belgium) \cite{wauters2015winning}, 
tackles the Swap-body Vehicle Routing problem. 
It is an iterated local search \cite{lourencco2010iterated} algorithm that uses late acceptance hill climbing \cite{burke2008late} as the local search component.
At each iteration of the late acceptance hill climbing, 
a neighborhood $N_k$ is randomly chosen from a large set of neighborhoods, 
and a neighbor solution $s’$ is generated according to $N_k$. 
The probability that a neighborhood $N_k$ is chosen is proportional to its weight value $w_k$. 
These weight values are fixed during each algorithm's run, and sum up to one. 
In addition, there are two integer parameters that control the late acceptance hill climbing: 
this local search component is stopped after a number of $itWI$ consecutive iterations without any improvement on the current solution, 
and the parameter $laList$ represents the size of the saved memory.

The algorithm consists of 42 neighborhoods, which were generated from 18 neighborhood types. 
Some of them are specially designed for the Swap-body Vehicle Routing problem (e.g., \lit{Convert-to-sub-route})
while the others are taken from the Vehicle Routing Problem literature (e.g., \lit{Cheapest-insertion}). 
Some neighborhood types can be parameterized by their sizes. 
For example: the size of a \lit{Cheapest-insertion} neighborhood is defined by the number of customers that will be removed and re-inserted back into the current solution.
We can have a small \lit{Cheapest-insertion} neighborhood with the size of 2, and a large \lit{Cheapest-insertion} neighborhood with the size of 25. 

Intuition can be used to reduce the number of weights to 28: some neighborhoods that belong to the same neighborhood type and have similar sizes can be grouped into one.
The list of neighborhood types and their groups of sizes are listed in Table \ref{tab:list of neighborhood types}. 
Parameter tuning is done on six (large) problem instances (\lit{large\_normal, large\_with, large\_without, new\_normal, new\_with, new\_without}) provided by the competition. An algorithm run on each instance takes 600 seconds. 
Note that the algorithm considered in this paper is actually not the same as the one that won the competition. 
The winning one is multi-threaded (4 independent parallel runs) while the one we use here is single-threaded. This is because the aim of our work is not to beat the winning algorithm, but to use this case study as a proof of concept for our characterization method.

\begin{table}
\caption{18 neighborhood types and 42 neighborhoods generated from them. Neighborhoods with sizes on the same line can be grouped into one to reduce the number of weight-value parameters to 28.}
\label{tab:list of neighborhood types}
\centering
\begin{center}
\begin{tabular}{|l|l|l|}
\hline
	Neighborhood type & Sizes \\ \hline
    \multirow{4}{*}{\lit{Cheapest-insertion}} & 1; 2; 3; 4; 5 \\ & 10;15 \\ & 20; 25 \\ & 35 \\ & 50 \\ \hline
    \lit{Swap} & \\ \hline
    \lit{Intra-route-two-opt} & \\ \hline
    \lit{Inter-route-two-opt} & \\ \hline
    \lit{Change-swap-location} & \\ \hline
    \lit{Merge-route} & \\ \hline
    \lit{Split-to-sub-routes} & \\ \hline
    \lit{Ruin-recreate} & 2; 3 \\ \hline
    \lit{Remove-route} & \\ \hline
    \lit{Remove-sub-route} & \\ \hline
    \lit{Remove-sub-route-with-cheapest-insertion} & \\ \hline
    \multirow{3}{*}{\lit{Remove-chain}} & 1; 2; 3; 4 \\ & 5; 6 \\ & 7; 8\\ \hline
    \multirow{3}{*}{\lit{Each-sequence-cheapest-insertion}} & (2,5) \\ & (5,2) \\ & (4,4) \\ \hline
    \lit{Convert-to-route} & \\ \hline
    \lit{Convert-to-sub-route} & \\ \hline
    \lit{Add-sub-route} & \\ \hline
    \multirow{4}{*}{\lit{Ejection-chain}} & 3; 4; 5 \\ & 10 \\ & 15 \\ & 35 \\ \hline
\hline
\end{tabular}
\end{center}
\end{table}

\section{Neighborhood characterization and clustering}
\label{sec:method}

Inspired by the idea from OSCAR \cite{mustafaoscar}, which is an automated approach for online selection of algorithm portfolio, 
we characterize each neighborhood $N_k$'s behaviours on an instance $I$ based on the following six observables: 

\begin{itemize}
\item Probabilities that $N_k$ improves, worsens or does nothing on a solution of $I$, denoted as $r_{improve}$, $r_{worsen}$, $r_{nothing}$, where $$r_{improve} + r_{worsen} + r_{nothing} = 1$$
\item Magnitudes of improvement and worsening, denoted as $a_{improve}$ and $a_{worsen}$
\item $N_k$'s running time (used for tie-breaking, as explained in section \ref{subsec: aggregate})
\end{itemize}

\begin{figure}
\centering
\includegraphics[height=6.2cm]{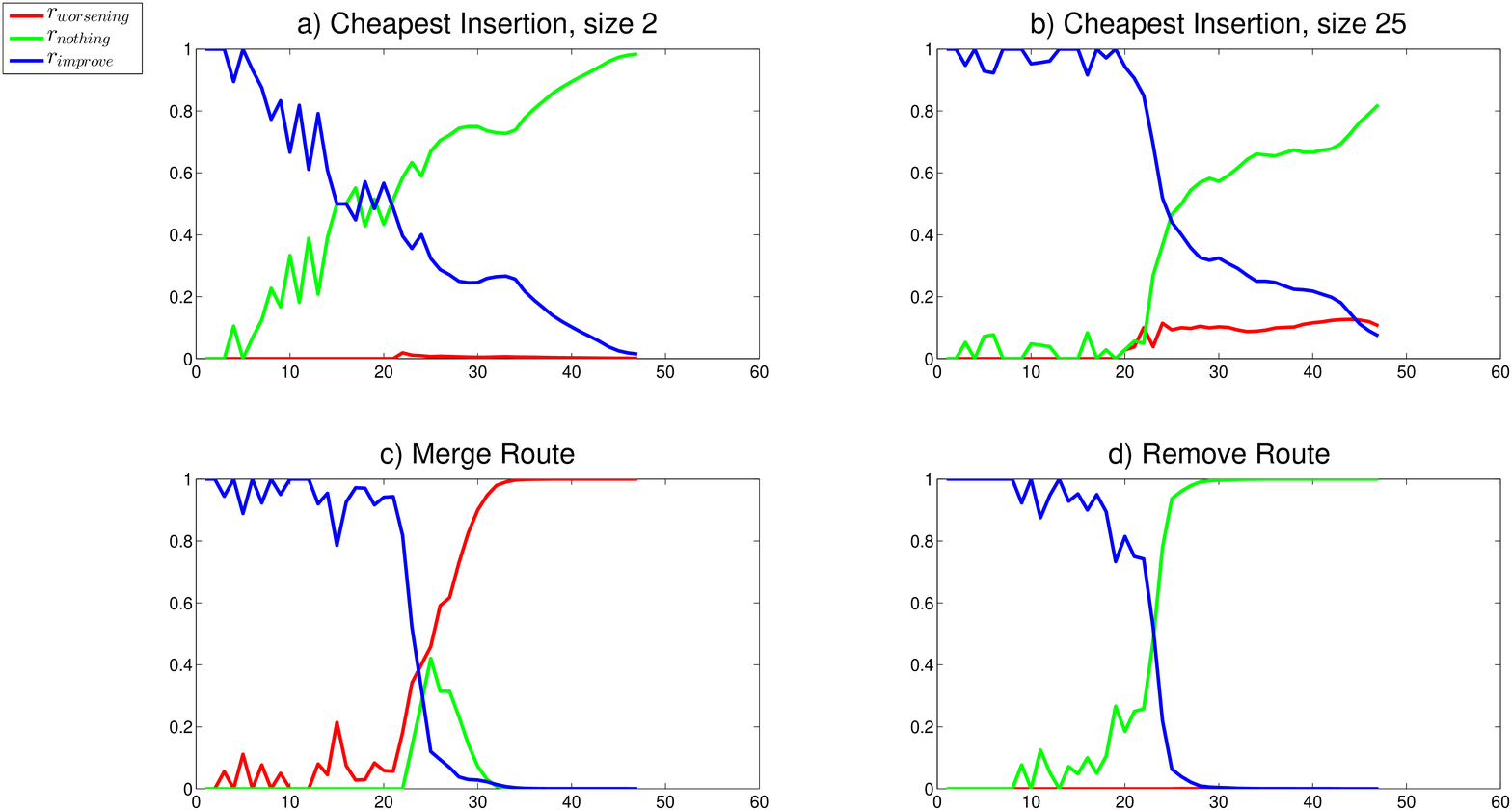}
\caption{Visualization of $r_{improve}$, $r_{worsen}$ and $r_{nothing}$ for four neighborhoods. The $x$-axis represents solution quality (the larger the value, the better the corresponding solution is). The $y$-axis represents values of the three observables.}
\label{fig:4 neighborhoods}
\end{figure}

The novelty of our method is that we will represent $N_k$ using the estimated values of those observables on different \textit{solution quality regions}, 
as they reflect changes of $N_k$'s behaviours according to the \textit{hardness} of the solution that it is dealing with. 
An illustration of such changes of $r_{improve}$, $r_{worsen}$ and $r_{nothing}$ for four neighborhoods on an instance is visualized in Figure \ref{fig:4 neighborhoods}. 
The $x$-axis represents solution quality (the larger the value, the better the corresponding solution is) and the $y$-axis represents values of the three observables. 
In order to draw those plots, we divide the range of solution quality into intervals, 
collect necessary information during algorithm runs, and group every ten intervals into one. Details on how to collect information for such a visualization will be described in section \ref{subsec:collect info}.
In Figure \ref{fig:4 neighborhoods}c, we can see that when the solution quality is low, i.e., the local search is in easy-to-improve region of the solution quality space, the \lit{Merge-route} neighborhood has a very high probability of improving the solution it is tackling. 
This probability drastically decreases when the neighborhood reaches a good solution quality region, and the probability of worsening the current solution starts reaching one from that point. 
On the other hand, the \lit{Remove-route} neighborhood in Figure \ref{fig:4 neighborhoods}d shows a similar behaviour in the low-solution quality region. 
However, in the good-solution quality region, the neighborhood tends to preserve the quality of the current solution rather than worsen it. 
Even neighborhoods belonging to the same neighborhood type can behave differently in different regions.
As shown in Figure \ref{fig:4 neighborhoods}a and \ref{fig:4 neighborhoods}b, the small \lit{Cheapest-insertion} neighborhood with size 2 has a much smaller probability of worsening a solution in the hard-to-improve region compared with the large \lit{Cheapest-insertion} neighborhood with size 25.

In the rest of this section, we introduce four steps to characterize and cluster neighborhoods. Firstly, necessary information is collected during algorithm’s runs. Then solution quality regions are automatically identified. Next, collected information on each region is aggregated to build neighborhoods’ feature vectors. Finally, we carry on cluster analysis.

\subsection{Collect necessary information during algorithm runs}
\label{subsec:collect info}
In this part, we describe the procedure of collecting all necessary information for characterizing neighborhood behaviours. Given a problem instance, we assume that an upper bound and a lower bound of the optimal solution quality are available. 
Since these bounds do not need to be tight, this assumption is not hard to be satisfied. 
For example, the upper bound could be obtained from a random solution or a solution generated by some greedy algorithm and the lower bound could result from solving a linear programming relaxation of the problem. 
In the algorithm considered in this work, the initial solution for each instance is produced by creating one route for each customer. 
We take that solution's value as the upper bound. A lower bound for each instance is provided by the authors of the algorithm, as the best solution obtained from running their best algorithm configuration (the multi-threaded version) in six hours. 

We divide the range between the upper bound and the lower bound into a large number of small intervals (here we set it as $1000$). Because higher quality solutions in general are harder to improve, we let the size of the intervals decrease exponentially. Each next interval has a size $0.99$ the size of the previous interval.

Now every time a neighborhood $N_k$ is applied on a solution that has quality value belonging to an interval $I_j$, the following values are accumulatively collected for the pair of ($N_k$, $I_j$): 

\begin{itemize}
\item $n_{iters}$: the number of times $N_k$ is applied,
\item $n_I$, $n_{SN}$, $n_W$: the numbers of times $N_k$ improves, does nothing, or worsens solutions, respectively,
\item $s_I$, $s_W$: sums of the amount of improvement and worsening,
\item $s_{time}$: sum of $N_k$'s running time.
\end{itemize}

Since the collection of these values is independent of algorithm configuration, 
it can be done during any algorithm runs, such as during testing, manual parameter tuning, or automated algorithm tuning. 
The more runs there are, the better the estimated values of the observables. 
In this work, we collect them from running two algorithm configurations on all instances, with 10 runs per instance, so the total number of algorithm runs is 240. 
We use a little bit longer running time (900 seconds per run) to make sure that the collected information can cover hard parts of the solution quality space.

\subsection{Identify solution quality regions as frames}
\label{subsec: identify regions}
\begin{figure}
\centering
\includegraphics[height=6.2cm]{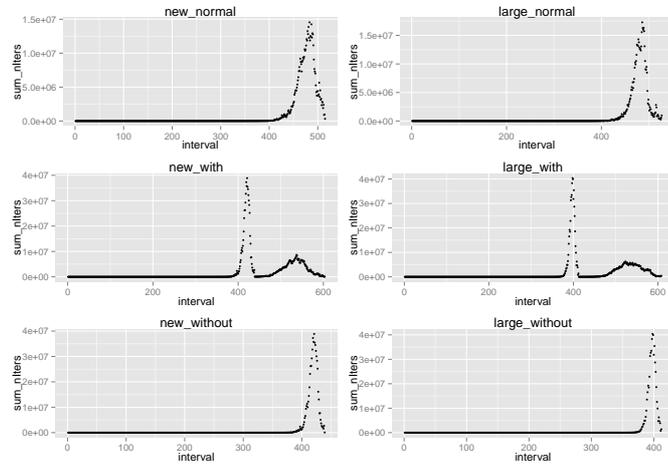}
\caption{$sum\_nIters$ on each instance. The $x$-axis represents solution intervals. The $y$-axis shows $sum\_nIters$.}
\label{fig:nIters_without_frameBounds}
\end{figure}

Intervals are grouped into frames based on $sum\_nIters$, the sum of all neighborhoods' $n_{iters}$ values on each interval. Figure \ref{fig:nIters_without_frameBounds} shows plots of $sum\_nIters$ on each instance. Note that because lower bounds on solution quality are not reached, intervals with zero $sum\_nIters$ at the end are removed. In this figure, there is a high peak in every plot, representing the interval where the algorithm stays most of the time. We thus conjecture that local optima or plateau should lie there. 
We can interpret the solution quality regions with low $sum\_nIters$ values before that peak as easy-to-reach and easy-to-escape, whereas regions around that peak as easy-to-reach and hard-to-escape and regions after that peak as hard-to-reach. 
The smaller peaks of two instances \lit{new\_with} and \lit{large\_with} should indicate second local optima or plateau. 
We propose Algorithm \ref{alg:group to frames} for grouping intervals into $n_{frames}$ regions (frames) that tries to reflect such an interpretation. 

\begin{algorithm}
\caption{Group intervals into frames}\label{alg:group to frames}
\begin{algorithmic}[1]
\Require $A$: the array of $sum\_nIters$ values, $n_{intervals}$: the number of intervals after removing empty ending intervals, $n_{frames}$: the number of frames
\Ensure $E$: an $n_{frames}$-element array, each element contains the index of the last interval of each frame
\State $r \leftarrow 0.05$, $avg \leftarrow \sum_{i=1}^{n_{intervals}}A[i]/n_{frames}$
\While{$true$}
	\State $E \leftarrow \emptyset, i \leftarrow 1, l \leftarrow avg * (1 + r)$ \label{alg: step 2}
    \While{$i \leq n_{interval}$}
		\If {$A[i] \geq l$}    	
	        \State $E \leftarrow E \cup \{i\}$, $i \leftarrow i + 1$
        \Else
        	\State \parbox[t]{\dimexpr\linewidth-\algorithmicindent}{Let $k$ be the largest value such that $\sum_{j=i}^{k}A[j] \leq l$\strut}
            \State $E \leftarrow E \cup \{k\}$
            \State $i \leftarrow k + 1$ 
        \EndIf
    \EndWhile
    \If {$E$ contains less than $n_{frames}$ elements}
    	\State $r \leftarrow r - 0.01$
    \Else
    	\If {$E$ contains more than $n_{frames}$ elements}
    		\State \parbox[t]{\dimexpr\linewidth-\algorithmicindent}{Starting from the first frame in $E$, combine every pair of frames into one (repeat from the first new frame if necessary) until only $n_{frames}$ elements are left.\strut}
        \EndIf
        \State break
    \EndIf
\EndWhile
\State Return $E$
\end{algorithmic}
\end{algorithm}

\begin{figure}
\centering
\includegraphics[height=6.2cm]{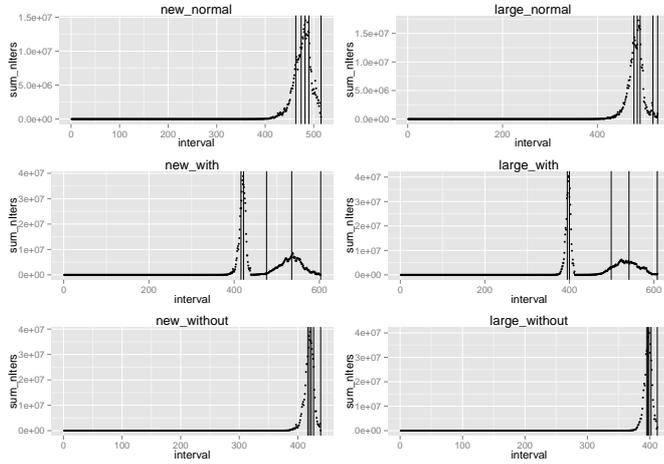}
\caption{$sum\_nIters$ on each instance, with frame boundaries shown as vertical lines for $n_{frames} = 5$. The $x$-axis represents solution intervals. The $y$-axis shows $sum\_nIters$.}
\label{fig:nIters_with_frameBounds}
\end{figure}

Figure \ref{fig:nIters_with_frameBounds} shows identified frames with $n_{frames} = 5$, which is used in our experiments, on the six provided instances.

\subsection{Characterize neighborhood behaviours as feature vectors by aggregating collected information into frames}
\label{subsec: aggregate}
For the first three observables, $r_{improve}$, $r_{worsen}$ and $r_{nothing}$,
we just simply sum the three values $n_I$, $n_W$ and $n_{SN}$ for all intervals belonging to the same frame. We then divide them by the sum of $n_{iters}$ to get the ratios. 
For the other two observables $a_{improve}$ and $a_{worsen}$, aggregation is more complicated. 
We can not sum $s_I$ or $s_W$ values over intervals and get the average due to the fact that their values are incomparable among different intervals. 
For example, we cannot say that an amount of improvement of $10$ in the two intervals $[33762, 33621)$ and $[33621,33482)$ are equal since hardness of solutions belonging to those is probably not the same. 
Therefore, we translate them into ranks before doing aggregation. 
For each interval, neighborhoods are ranked based on the averages of their corresponding $s_I$, $s_W$ values. 
Because ties can happen, e.g., some neighborhoods might never make any improvement in the hard solution quality regions, the average value of $s_{time}$ in the corresponding interval is used for tie-breaking. Since the intervals are fine, the resulting ranked lists are possibly:
\begin{itemize}
\item noisy: at some intervals, $n_{iters}$ of some neighborhoods might be very small, so that their estimated values of $a_{improve}$ and $a_{worsen}$ might be inaccurate. 
\item partial: $n_{iters}$ of some neighborhoods might be equal to zero at some intervals, i.e., we do not have information of those neighborhoods in those intervals. 
\end{itemize} 
Therefore, we aggregate them using the R package \textit{RobustRankAggreg} - a robust ranking aggregation method \cite{kolde2012robust} specially designed for similar situations in bioinformatics. 
Eventually, for each neighborhood, we have a feature vector composing of 150 components, which is a combination of 5 observables, 5 frames and 6 instances.

\subsection{Cluster analysis on neighborhoods}
\label{subsec: cluster analysis}
The first three observables, $r_{improve}$, $r_{worsen}$ and $r_{nothing}$, sum up to one. As a result, their corresponding vector components belong to a special class named \textit{compositional data}. 
As explained in \cite{Aitchison2005} ``sample space for compositional vectors
is radically different from the real Euclidean space associated with
unconstrained data'', multivariate statistical methodology designed for unconstrained data could not be applied directly. To convert them back to the Euclidean space, we apply the isometric log-ratio transformation proposed in \cite{egozcue2003isometric}. 
After the transformation, since the three observables are reduced to two, each feature vector is now 120-dimensional. 
We can start doing cluster analysis on neighborhoods based on those vectors. 
Since the number of dimensions ($120$) is larger than the number of individuals ($42$), the clustering method High-Dimensional Data Clustering
(HDDC) \cite{bouveyron2007high}), which is implemented in the R package HDclassif
\cite{berge2012hdclassif}, is used for cluster analysis. 
This method has two desirable properties: the ability of dealing with high-dimensional 
low-sample data, and the optimal number of clusters automatically decided 
based on Bayesian Information Criterion. In the end, 42 neighborhoods are grouped into 9 clusters:
\begin{itemize}
\item \lit{Ejection-chain 3, 4, 5; Remove-chain 1, 2, 3, 6, 7, 8; Remove-sub-route-with-cheapest-insertion;
\item Swap; Inter-route-two-opt
\item Cheapest-insertion 10, 15, 20, 25, 35, 50; Each-sequence-cheapest-insertion (2,5), (4,4), (5,2); Remove-chain 4
\item Cheapest-insertion 1, 2, 3, 4, 5
\item Change-swap-location; Merge-route
\item Add-sub-route; Convert-to-sub-route
\item Ejection-chain 10, 15, 35; Remove-chain 5; Intra-route-two-opt
\item Ruin-recreate 2, 3
\item Convert-to-route; Remove-sub-route; Remove-route; Split-to-sub-route}
\end{itemize}
It might be interesting to have a look at some of the resulting clusters. The two neighborhoods \lit{Merge-route} and \lit{Remove-route} behaves quite differently in the second-half region as shown in Figure \ref{fig:4 neighborhoods}, and they are indeed clustered into two different groups. By taking a look into the neighborhoods' implementation, we know that \lit{Add-sub-route} and \lit{Convert-to-sub-route} have an extreme behaviour when compared to the others: they will add an additional cost into the current solution and worsen it most of the time (it can also be seen in plots of their observables, which are similar to the ones shown in Figure \ref{fig:4 neighborhoods}). We can say that the cluster analysis does recognize this extremeness, as the two neighborhoods are grouped into a separated cluster. In addition to reflecting knowledge that can be guessed by looking at the neighborhoods' implementation, the cluster analysis also does some grouping that is not intuitive from the neighborhoods' structure, e.g., the grouping of \lit{Ejection-chain 10,15,35} and \lit{Intra-route-two-opt}.

\section{Experimental results}
\begin{figure}
\centering
\includegraphics[height=6.2cm]{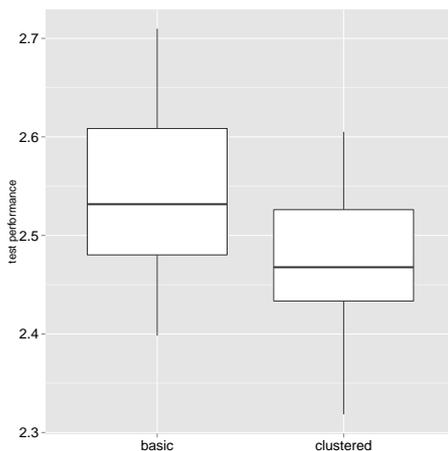}
\caption{Test performance of the two considered scenarios}
\label{fig:13_14}
\end{figure}

Our hypothesis is that the proposed characterization method does reflect neighborhood behaviours on the given set of instances, so that the generated feature vectors should correctly represent the neighborhoods and the clusters we obtained are meaningful. To test this hypothesis, we applied the automated tuning tool
SMAC \cite{hutter2011sequential} to two configuration scenarios: the first one, dubbed \lit{basic}, uses the 28 groups of neighborhoods described in Table \ref{tab:list of neighborhood types}, the second one, dubbed \lit{clustered} uses the 9 clusters of neighborhoods generated from our characterization method. We carried out 18 runs of SMAC on each scenario. Each one has a budget of 2000 algorithm runs (13.9 CPU days). Due to the large CPU time each SMAC run requires, we use the shared-model-mode offered by SMAC with 10 cores (walltime is 1.39 days), and take the configuration which has the best training performance as the final one. 
Mean of optimality gaps (in percentage) on the instance set is used as tuning performance measure. Optimality gap on each instance is calculated by: $$ optimalityGap = 100 * (solutionCost - lowerBound) / lowerBound$$
where $lowerBound$ is provided by the algorithm's authors, and is the best solution cost obtained after running the multi-threaded version of the algorithm on the corresponding instance in 6 hours.
The best algorithm configuration from each SMAC run is evaluated using test performance, which is the mean of optimality gaps obtained from 30 runs of the configuration on the instance set (5 runs/instance). Box-plots of the 18 SMAC runs on each scenario are shown in Figure \ref{fig:13_14}, in which the \lit{clustered} scenario offers advantage over the \lit{basic} scenario. A \textit{paired t-test} is conducted and gives a $p-value$ of $0.009258918$, indicating statistical significance.

\begin{figure}
\label{fig:13-14-13_weight1-14_weight1}
\centering
\includegraphics[height=6.2cm]{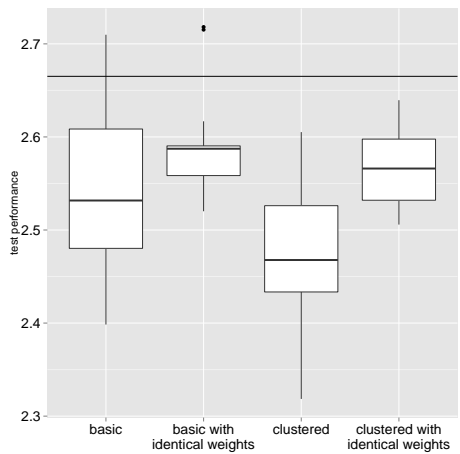}
\caption{Test performance of the two considered scenarios, their SR versions and the default configuration}
\end{figure}

In the hyper-heuristic community, in particular the Selection Hyper-heuristic class, in which the aim is to manage a set of low-level heuristics during the search by selecting one of them at each iteration, the Simple Random (SR) heuristic selection mechanism is often used as a baseline \cite{burke2013hyper}. In our setting, SR is equivalent to the parameter configuration that has identical weights for all neighborhoods. It will be interesting to compare SR with the resulting configurations obtained from the off-line tuning: for each scenario, the 18 best tuned configurations are taken and the neighborhood weights inside them are set to identical. Their test performance values are shown as \textit{basic with identical weights} and \textit{clustered with identical weights} in Figure \ref{fig:13-14-13_weight1-14_weight1}. The horizontal line represents test performance of the algorithm configuration in which neighborhood weights are identical and \textit{laList} and \textit{itIW} are set to values recommended by the algorithm's authors. This configuration is also used as the default configuration for all SMAC runs mentioned above. We can see that the SR versions in both scenarios give worst test performance. A paired \textit{t-test} is conducted for each scenario:
\begin{itemize}
\item \textit{basic} and \textit{basic with identical weights}: \textit{p-value = 0.07464}
\item \textit{clustered} and \textit{clustered with identical weights}: \textit{p-value = 0.000459}
\end{itemize}
The \textit{p-value} from the second \textit{t-test} indicates that the neighborhoods' weights do have influence on the algorithm performance. Those tests also reflect the hardness of tuning those weights (as the \textit{basic} tuning fails to show significant improvement over the identical-weight configurations), and the advantage of \textit{clustered} over \textit{basic}.

\section{Conclusion and future work}
In this paper, we have proposed a systematic method to characterize neighborhood behaviours in a multi-neighborhood local search framework, where the probability of choosing a neighborhood at each iteration is chosen in an off-line manner. The characterization is based on the probabilities that a neighborhood will improve, worsen or do nothing on a solution, on the magnitudes of its improvement and worsening, and on its running time. We have observed that these characteristics change according to hardness of different regions in solution quality space. As a result, we design our method such that it tries to detect these regions based on collected information and represent neighborhood behaviours on them as feature vectors. Cluster analysis is then applied to form groups of similar neighborhoods. A tuning experiment with the automated algorithm configuration tool SMAC \cite{hutter2011sequential} shows that using these clusters gives a statistically significant improvement on test performances of the obtained algorithm configurations over the non-clustered version. It verifies the hypothesis that our characterization method is able to correctly reflect neighborhood behaviours on the given instance set. Since the information used in this method does not depend on a specific problem, the characterization and clustering procedure potentially can be applied in similar contexts. A first version of our method's implementation has been available as a toolbox, and can be obtained by sending a request to the corresponding author. The toolbox receives log files containing necessary information collected during algorithm runs as input, and returns results of the cluster analysis, as well as box-plots and graphs for the visualization of observables and solution quality regions.

For future work, a multi-level tuning might be interesting. Firstly, a post-analysis on the importance of each cluster using fANOVA \cite{hutter2014efficient}, which is an efficient approach to ``quantify the effect of algorithm parameters'', can be applied. Then finer tuning on neighborhoods that belong to the most important clusters can be done. In addition, since our current method are only limited to a small number of instances, we are seeking for the possibility of an extension to a large set of instances. We might want to exploit problem-specific expert knowledge, e.g., instance features, in such a case.

\section*{Acknowledgement}
This work is funded by COMEX (Project P7/36), a BELSPO/IAP Programme. We would like to thank T\'{u}lio Toffolo for his great support during the course of this research, Thomas St\"utzle and Jan Verwaeren for their valuable remarks. The computational resources and services used in this work were provided by the VSC (Flemish Supercomputer Center), funded by the Hercules Foundation and the Flemish Government – department EWI. 

\bibliography{references}
\bibliographystyle{unsrt}
\end{document}